\documentclass[lettersize,journal]{IEEEtran}
\usepackage{amsmath,amsfonts}
\usepackage{algorithmic}
\usepackage{algorithm}
\usepackage{array}
\usepackage[caption=false,font=normalsize,labelfont=sf,textfont=sf]{subfig}
\usepackage{textcomp}
\usepackage{stfloats}
\usepackage{url}
\usepackage{verbatim}
\usepackage{graphicx}
\usepackage{cite}
\usepackage{multirow}
\usepackage{booktabs} 
\usepackage{colortbl}
\usepackage{hyperref}
\usepackage{arydshln}
\usepackage{pifont}
\newcommand{\xmark}{\ding{55}} 
\newcommand{\cmark}{\ding{51}} 
\usepackage[table]{xcolor} 

\begin{document}

\title{EyeSim-VQA: A Free-Energy-Guided Eye Simulation Framework for Video Quality Assessment}

\author{Zhaoyang Wang, Wen Lu, \textit{Member, IEEE,} Jie Li, Lihuo He, \textit{Member, IEEE,} Maoguo Gong, \textit{Fellow, IEEE,} Xinbo Gao, \textit{Fellow, IEEE} 
\thanks{
This work was supported in part by the National Natural Science Foundation of China under Grants 62036007, 62101084, 62221005, 62171340, 62476207; in part by the Chongqing Natural Science Foundation Innovation and Development Joint Fund Project under Grants CSTB2023NSCQ-LZX0085 and CSTB2023NSCQ-BHX0187; in part by the Key Industrial Innovation Chain Project in Industrial Domain of Shaanxi Province under Grant No. 2020ZDLGY05-01; in part by the Fundamental Research Funds for the Central Universities under Grant No.  YJSJ25008. (Corresponding authors: Xinbo Gao.)

Zhaoyang Wang, Wen Lu, Jie Li and Lihuo He are with the State Key Laboratory of Integrated Services Networks, School of Electronic Engineering, Xidian University, Xi’an, Shaanxi 710071, China (e-mail: zywang23@stu.xidian.edu.cn; luwen@mail.xidian.edu.cn; leejie@mail.xidian.edu.cn and lhhe@mail.xidian.edu.cn). 

Maoguo Gong is with the Key Laboratory of Collaborative Intelligence Systems, Ministry of Education, Xidian University, Xi’an 710071, China, and is also affiliated with the College of Mathematical Science, Inner Mongolia Normal University, Hohhot 010028, China (e-mail: gong@ieee.org).

Xinbo Gao is with the School of Electronic Engineering, Xidian University, Xi’an 710071, China (e-mail: xbgao@mail.xidian.edu.cn).
}
}

\markboth{Journal of \LaTeX\ Class Files,~Vol.~14, No.~8, August~2021}%
{Shell \MakeLowercase{\textit{et al.}}: A Sample Article Using IEEEtran.cls for IEEE Journals}


\maketitle

\begin{abstract}

Free-energy-guided self-repair mechanisms have shown promising results in image quality assessment (IQA), but remain under-explored in video quality assessment (VQA), where temporal dynamics and model constraints pose unique challenges. Unlike static images, video content exhibits richer spatiotemporal complexity, making perceptual restoration more difficult. Moreover, VQA systems often rely on pre-trained backbones, which limits the direct integration of enhancement modules without affecting model stability.
To address these issues, we propose EyeSimVQA, a novel VQA framework that incorporates free-energy-based self-repair. It adopts a dual-branch architecture, with an aesthetic branch for global perceptual evaluation and a technical branch for fine-grained structural and semantic analysis. Each branch integrates specialized enhancement modules tailored to distinct visual inputs—resized full-frame images and patch-based fragments—to simulate adaptive repair behaviors. We also explore a principled strategy for incorporating high-level visual features without disrupting the original backbone. In addition, we design a biologically inspired prediction head that models sweeping gaze dynamics to better fuse global and local representations for quality prediction.
Experiments on five public VQA benchmarks demonstrate that EyeSimVQA achieves competitive or superior performance compared to state-of-the-art methods, while offering improved interpretability through its biologically grounded design.

\end{abstract}


\section{Introduction}
\IEEEPARstart{W}{ith} the rapid advancement of digital technology, the barrier to generating video content has significantly decreased, leading to a surge in user-generated content (UGC) across various platforms such as YouTube and TikTok. However, due to the heterogeneous nature of filming equipment, user expertise, diverse content styles and bitrate constrain \cite{wu2023discovqa,liu2021spatiotemporal,qi2025generative,dong2023temporal,fang2020perceptual,guo2023learning}, UGC videos often exhibit large variations in visual quality. Accurately assessing and predicting the perceptual quality of UGC videos remains a long-standing and important research challenge in the field of video understanding.
To support the task of video quality assessment (VQA), numerous datasets \cite{hosu2017konstanz,wang2019youtube,ying2021patch,lu2024kvq,duan2024finevq,sinno2018large} have been developed in recent years. In particular, several datasets target content from short video platforms (e.g., S-UGC, KVQ \cite{lu2024kvq}), while others focus on multi-dimensional quality annotations (e.g., FineVD \cite{duan2024finevq}). The continued refinement and expansion of such datasets play a crucial role in enhancing the robustness and generalizability of VQA algorithms.

Early VQA methods \cite{hassen2013image,kundu2017no,mittal2012no,mittal2012making,xue2014blind} primarily relied on hand-crafted features to model spatial and temporal distortions. While these approaches are lightweight and computationally efficient, they often fail to capture the complex perceptual cues inherent in UGC videos. As a result, their predictions frequently deviate from human subjective quality judgments, leading to suboptimal performance.
With the rapid advancement of deep learning, recent VQA methods \cite{li2019quality,wu2022fast,wu2023discovqa,wu2023neighbourhood,wu2022disentangling,lu2024kvq,duan2024finevq,mi2024clif} based on deep neural networks have demonstrated significantly improved capabilities in addressing the diverse and unstructured nature of UGC videos. These models can automatically learn hierarchical and abstract visual representations from raw data, enabling more accurate and perceptually aligned quality predictions compared to traditional feature-based approaches.

The evolution of VQA methods—from traditional hand-crafted approaches to deep learning-based techniques and their iterative refinement—reflects a continuous effort to approximate human subjective perception of video quality. Early methods relied on manually designed features, while modern approaches leverage neural networks to extract high-level, abstract representations that align more closely with human visual experience.
The transition from convolutional neural network (CNN)-based models \cite{hara2017learning,hara2018can,he2016deep,he2016identity,tan2019efficientnet,tan2021efficientnetv2} to Transformer-based architectures \cite{wu2023discovqa,wu2022fast,wu2023neighbourhood,lu2024kvq} further enhances the ability to capture long-range dependencies and global context. For instance, FastVQA \cite{wu2022fast} preserves patch-level information to better model high-resolution details, simulating how humans perceive fine-grained objects within a visual field. The Dover \cite{wu2022disentangling} explicitly integrates both local details and global semantics, mirroring the dual-level nature of human visual assessment.
More recently, the incorporation of large language models (LLMs) \cite{lu2024kvq,duan2024finevq,mi2024clif} introduces semantic-level reasoning into the VQA pipeline, aligning even more closely with the way humans interpret and evaluate video content. These trends collectively highlight a dominant paradigm in VQA research: \textit{\textbf{designing models that are better aligned with human subjective visual perception}} \cite{wu2022disentangling,mi2024clif}.


Despite the recent progress in VQA, approaches inspired by the free energy principle—particularly those modeling the human eye’s self-repair mechanism \cite{pan2022vcrnet,wang2025diffusion}, wherein the visual system adaptively reconstructs incomplete or degraded content—\textit{\textbf{remain largely unexplored in this domain}}. \textbf{\textit{Such mechanisms are highly consistent with how humans perceive and interpret visual stimuli in natural viewing scenarios.}} Notably, free-energy-guided methods have already demonstrated promising results in the field of image quality assessment (IQA) \cite{ren2018ran4iqa,lin2018hallucinated}, suggesting their potential effectiveness when extended to the more complex setting of VQA.

However, effectively extending the free-energy-based self-repair mechanism to the VQA domain presents several open challenges. Unlike static images, video data is inherently more complex, containing rich spatiotemporal information and dynamic content \cite{chan2021basicvsr,chan2022basicvsr++}. Moreover, VQA backbone networks are typically deeper and more computationally intensive than those used in IQA \cite{chen2021learning,liu2018end,wu2022disentangling,lu2024kvq}, making integration more difficult. In this context, designing models that can simulate the human visual system’s self-repair behavior in a temporally coherent manner becomes a non-trivial task. Another critical challenge lies in how to incorporate high-level perceptual features, guided by the free energy principle, into existing VQA architectures without disrupting their structure or efficiency
\cite{wu2022disentangling,lu2024kvq,liu2018end}. Additionally, it is essential to explore strategies that align more closely with the subjective visual evaluation process \cite{wu2022disentangling}, capturing both low-level and high-level details in a human-consistent manner.

In this paper, to address the aforementioned challenges, we propose a free-energy-guided \textbf{Eye}
\textbf{Sim}ulation framework for \textbf{VQA} (EyeSim-VQA), a novel VQA model inspired by the free-energy-guided self-repair mechanism of the human visual system. This design is aligned with the recent research trend of constructing models that better reflect human subjective visual perception.
Specifically, we design a dual-branch architecture for visual enhancement and quality evaluation. One branch operates on the entire video frame to perform global-level enhancement and quality regression, while the other branch processes combinations of patch fragments, retaining local detail features for localized enhancement and re-evaluation.
To accommodate different types of video features, we construct tailored visual enhancement networks that more effectively simulate the human eye's adaptive repair mechanism. Furthermore, we conduct experiments to explore the integration of high-level perceptual features into the VQA pipeline.
In addition, to further align with human visual behavior, we design a two-branch prediction head based on DyT \cite{zhu2025transformers} and Mamba \cite{gu2023mamba} architectures, simulating the human visual system's scanning and fixation behaviors. The overall architecture jointly models global-local perception, the free-energy-based self-repair process, and scan-and-gaze dynamics, aiming to approximate real human visual experience during VQA.

Our contributions can be summarized as follows:
\begin{enumerate}
    \item To the best of our knowledge, this work makes the first attempt to introduce the human eye’s self-repair mechanism, guided by the free energy principle, into the VQA domain. We propose EyeSim-VQA, a model that more closely mimics the human visual system when evaluating video quality.
    \item We design a dual-branch prediction head that simulates the human eye’s scan-and-gaze mechanism by integrating DyT and Mamba architectures, enabling the model to capture both global and local perceptual cues. This structure closely aligns with human visual behavior and explores the potential of these architectures in the VQA domain.
    \item Extensive experiments on multiple public VQA datasets demonstrate the effectiveness of the proposed method. EyeSim-VQA achieves state-of-the-art (SOTA) performance compared to existing mainstream approaches.
\end{enumerate}

The remainder of this paper is organized as follows. Section \ref{s2} reviews related work on VQA and free-energy-guided visual modeling. Section \ref{s3} presents the proposed method in detail. Section \ref{s4} reports the experimental results. Section \ref{s5} discusses the limitations of our approach and potential future directions. Section \ref{s6} concludes the paper.

\section{Related Work} \label{s2}
\subsection{Classical VQA Methods}
Traditional VQA methods mainly rely on hand-crafted features to characterize spatial and temporal degradations in video content for quality evaluation. Since a video consists of a sequence of image frames, early VQA approaches typically applied IQA techniques to individual frames \cite{hassen2013image,kundu2017no,mittal2012no,mittal2012making,xue2014blind}. For instance, V-CORNIA \cite{xu2014no} extends the original CORNIA \cite{ye2012unsupervised} model—originally designed for IQA—to the video domain. While such methods are effective at assessing the quality of individual frames, they fail to capture temporal consistency and coherence across frames, which significantly limits their overall performance in VQA tasks.
Subsequent methods based on natural video statistics (NVS) have been proposed to jointly capture both spatial and temporal information in video content. V-BLIINDS \cite{saad2014blind}, for example, integrates DCT-domain analysis with the NIQE \cite{mittal2012making} framework to extract inter-frame variations for quality prediction. TLVQM \cite{korhonen2019two} introduces a two-level feature extraction strategy, capturing inter-frame motion and action-related features along with spatial features from individual frames, leading to a more comprehensive assessment. VIDEVAL \cite{tu2021ugc} combines multiple hand-crafted features to represent complex video characteristics while significantly reducing the overall feature complexity, making it efficient and interpretable.

Although classical VQA methods are computationally efficient and easy to implement, they struggle to deliver satisfactory performance when applied to increasingly complex and diverse UGC videos.

\subsection{Deep Learning-based VQA Methods}
Deep learning-based VQA methods have gradually surpassed traditional approaches, primarily due to their ability to adaptively extract high-level and abstract visual features. These methods typically adopt a backbone network for feature extraction, followed by a VQA prediction head to generate the final Mean Opinion Score (MOS).
Early deep learning VQA models were predominantly CNN-based \cite{hara2017learning,hara2018can,he2016deep,he2016identity,tan2019efficientnet,tan2021efficientnetv2}. For example, VSFA \cite{li2019quality} employs ResNet-50 \cite{he2016deep}, pre-trained on ImageNet \cite{deng2009imagenet}, to extract spatial features and utilizes a GRU \cite{cho2014learning} module to capture temporal dependencies. Similarly, GSTVQA \cite{chen2021learning} uses VGG-16 as its feature extractor. With the advancement of CNN architectures, 3D-CNNs were introduced to jointly model spatial and temporal information. For example, V-MEON \cite{liu2018end} applies 3D-CNNs to extract spatiotemporal features in a unified manner and also explores a hybrid framework that combines 2D-CNNs and 3D-CNNs for more effective feature learning.
More recently, inspired by the superior performance of Transformer architectures across various vision tasks, Transformer-based VQA models \cite{wu2023discovqa,wu2022fast,wu2023neighbourhood,lu2024kvq} have emerged. DisCoVQA \cite{wu2023discovqa}, for instance, employs the Video Swin Transformer \cite{liu2022video} as a backbone to better capture spatiotemporal dependencies, resulting in improved accuracy and generalization across diverse video content.
Although these methods have achieved promising results, most of them primarily benefit from advancements in network architectures, rather than explicitly incorporating modeling principles grounded in human subjective visual perception.

In recent years, several methods have been proposed to better align with the subjective visual perception of human observers, achieving notable improvements in performance. FastVQA \cite{wu2022fast} and FasterVQA \cite{wu2023neighbourhood} introduce a fragment-based evaluation strategy by losslessly cropping and stitching original video frames, as opposed to the conventional resizing approach. This preserves fine-grained visual details and mimics the subtle observation process of the human visual system.
Building on this idea, the Dover \cite{wu2022disentangling} model employs a dual-branch architecture: one branch focuses on fine-grained fragment-level evaluation, while the other performs holistic assessment on the entire frame. This design is consistent with the way humans combine global judgment with attention to local details during video observation.
More recent approaches incorporate semantic information to further enhance perceptual alignment \cite{lu2024kvq,duan2024finevq,mi2024clif}. For example, KVQ \cite{lu2024kvq} and CLiF-VQA \cite{mi2024clif} utilize the CLIP model to extract semantic-level features, which are then used as intermediate representations in conventional VQA pipelines where the final quality score is predicted through a VQA head.
In contrast, FineVQ \cite{duan2024finevq} represents a shift toward fully LLM-based, end-to-end VQA architectures. It directly feeds video content into a vision-language model and predicts quality scores across multiple dimensions, offering a holistic and semantically enriched evaluation that more closely mirrors human subjective assessment.

These trends indicate that designing VQA models aligned with human subjective visual perception is a promising direction for addressing the challenges of UGC-VQA. Motivated by this, we introduce the free-energy-guided self-repair mechanism of the human visual system into the VQA framework, aiming to simulate the perceptual process of human observers more effectively.

\subsection{Related Free-Energy-Guided IQA Methods}
Inspired by the free-energy principle, which posits that the human visual system (HVS) can automatically restore and enhance distorted visual inputs, researchers have proposed no-reference image quality assessment (NR-IQA) methods guided by this principle. These approaches typically model the quality reconstruction relationship between a distorted image and its perceptually restored version, aiming to approximate the HVS’s self-repair mechanism during quality evaluation.
In \cite{ren2018ran4iqa}, Ren \textit{et al.} propose the RAN4IQA model, which employs a restorer-discriminator architecture to jointly recover distorted images and evaluate quality based on both restored and original inputs. In \cite{lin2018hallucinated}, Li \textit{et al.} utilize adversarial learning to generate hallucinated reference images, serving as high-level compensatory information for quality prediction. In \cite{pan2022vcrnet}, the instability of adversarial training is addressed by designing a dedicated restoration network, whose intermediate feature representations are used to guide the main IQA model. To overcome the lack of interpretability in these intermediate variables, in \cite{wang2025diffusion}, Wang \textit{et al.} introduce diffusion models into the IQA domain for the first time. By leveraging the intermediate denoising stages, the model generates explicit perceptual cues to assist the quality estimation process.

These methods demonstrate the effectiveness of the free-energy-guided self-repair mechanism in IQA, and also suggest its promising potential for extension to the more complex task of VQA.

\begin{figure*}[t]
    \centering
    \includegraphics[width=1\linewidth]{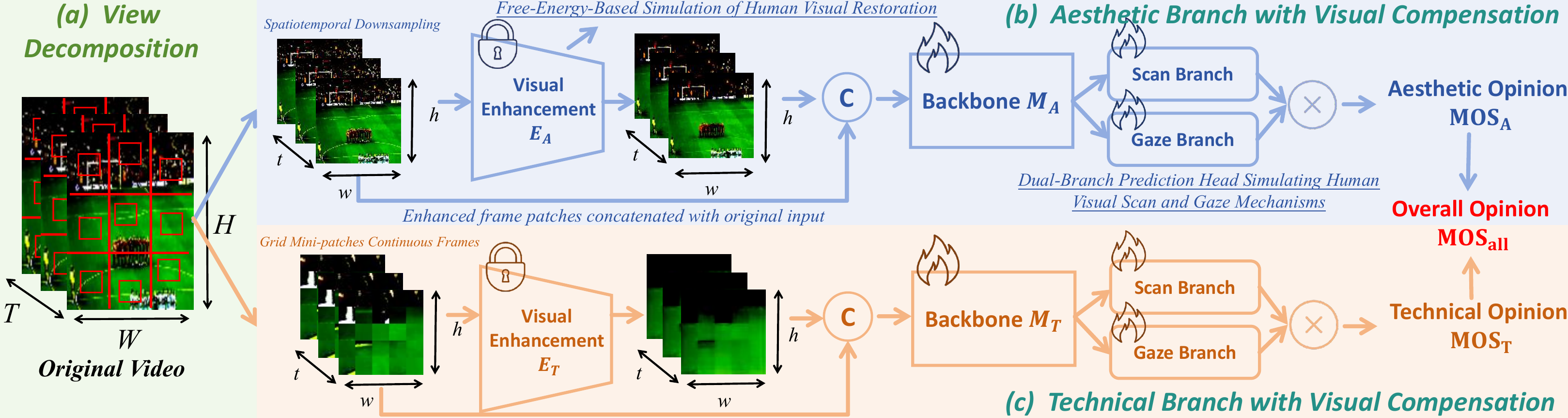}
    \caption{Overview of the proposed EyeSim-VQA framework.
(a) View Decomposition module, shared with existing models such as DOVER \cite{wu2022disentangling} and KSVQE \cite{lu2024kvq}.
(b) Aesthetic branch and (c) Technical branch, both simulate human visual restoration via a vision enhancement network applied to the original image. The enhanced features are then fed into the different backbones for feature extraction, followed by a dual-branch VQA prediction head. Final MOS is obtained by fusing predictions from both branches.}
    \label{main}
\end{figure*}

\section{Methodology} \label{s3}

\subsection{Pipeline of the proposed EyeSim-VQA}
The overall architecture of our model is illustrated in Fig. \ref{main}. Given an input video, we first perform view decomposition tailored to two parallel branches. Specifically, the video is resized to a lower resolution in the aesthetic branch to capture holistic appearance features, while in the technical branch, it is partitioned into patch-based chunks to preserve fine-grained structural details.
To simulate the human eye’s self-repair mechanism, each branch incorporates a dedicated visual enhancement network that improves perceptual quality before feature extraction. The enhanced outputs are concatenated with the original inputs and passed through the corresponding backbone networks to extract refined high-level visual representations.
These representations are then fed into our dual-branch prediction head, which integrates two biologically inspired mechanisms. A Mamba-based scanning module \cite{gu2023mamba,liu2024vision,zhang2024survey} is used to mimic the sequential visual exploration behavior of human perception, while a attention-based gaze module simulates focused visual attention. Finally, the outputs from both branches are fused to produce the final predicted quality score.

The following sections elaborate on the core components of our framework. Section \ref{sec_ve} introduces the visual enhancement module, which simulates the human eye's self-repair capability to improve perceptual quality. Section \ref{sec_eyesim} presents the EyeSim-VQA prediction head that models human visual perception via dual mechanisms—gaze and scanning—corresponding to two functional branches. Section \ref{sec_loss} details the design of the loss function used to supervise the visual restoration process.

\begin{figure*}[t]
    \centering
    \includegraphics[width=0.66\linewidth]{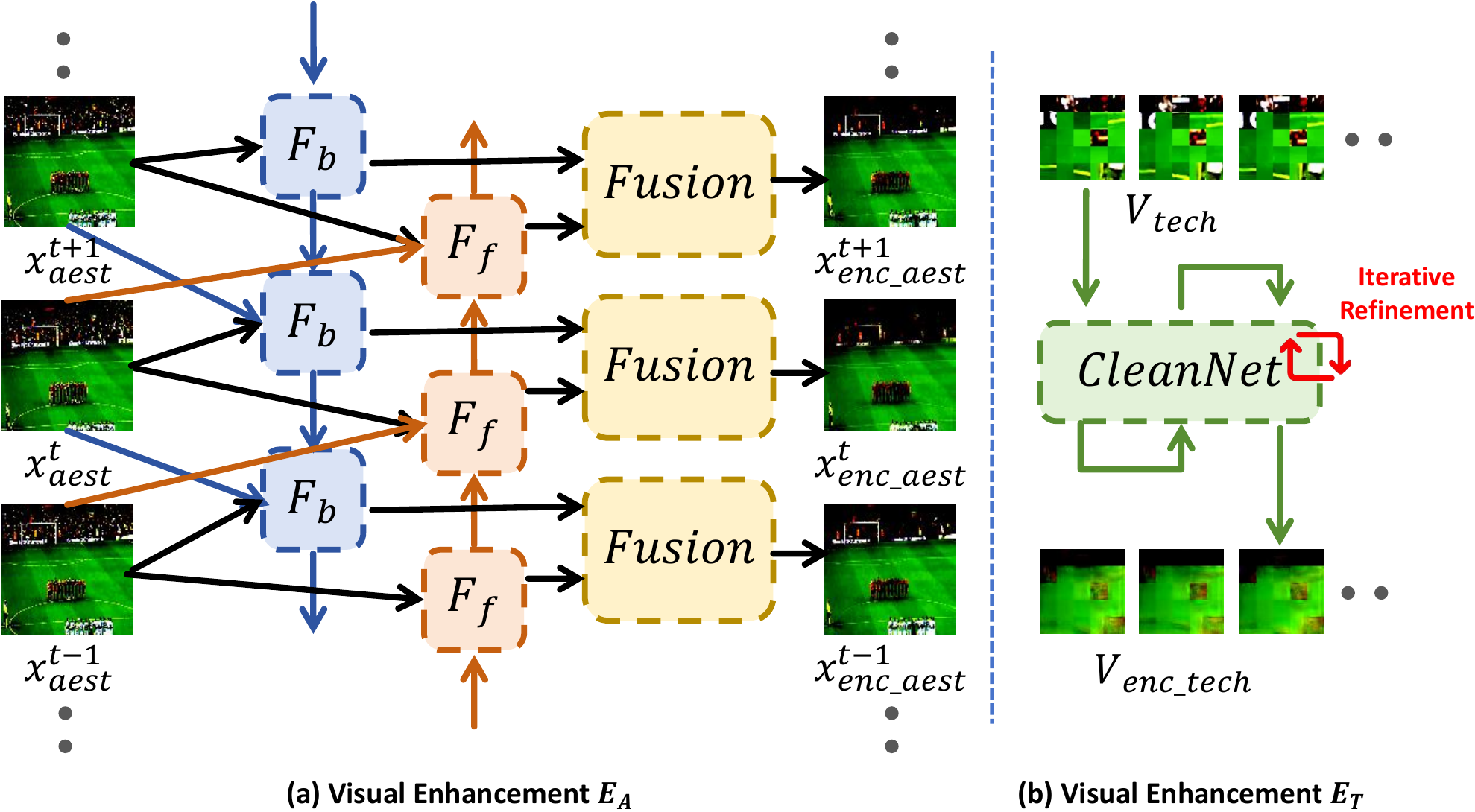}
    \caption{Illustration of the visual enhancement module. (a) The aesthetic branch adopts a lightweight BasicVSR \cite{chan2021basicvsr} (BasicVSR-mini) architecture with reduced parameter count and fewer residual blocks, where $F_b$ and $F_f$ denote backward and forward propagation, respectively. Each propagation includes three key components: a flow estimation module, a spatial warping module, and residual blocks. (b) The technical branch employs CleanNet, which consists of residual blocks and uses iterative refinement loops for video quality restoration.}
    \label{E_A_and_E_T}
\end{figure*}

\subsection{Visual Enhancement via Human Eye-Inspired Self-Repair} \label{sec_ve}

\subsubsection{\textbf{Visual Enhancement in the Aesthetic Branch}}
In the aesthetic branch, we aim to evaluate the global perceptual quality of a video by operating on uniformly resized frames sampled at fixed intervals. This strategy preserves inter-frame variations while reducing computational overhead. To enhance visual quality prior to assessment, we adopt BasicVSR \cite{chan2021basicvsr}—a super-resolution model based on optical flow and pixel-level processing—as the foundation for our visual enhancement module. Leveraging temporal information from adjacent frames, the model effectively captures inter-frame differences, while the use of optical flow ensures accurate spatial alignment across frames. These properties make BasicVSR particularly suitable for holistic enhancement in the aesthetic branch.

However, since our objective is not full-resolution reconstruction but rather perceptual refinement and noise suppression, we prioritize efficiency over complexity. To this end, we introduce a lightweight variant, BasicVSR-mini, as illustrated in Fig.~\ref{E_A_and_E_T}(a). Specifically, we significantly reduce the number of residual blocks and intermediate feature channels, and remove all components related to spatial resolution changes from the original architecture. This streamlined design effectively simulates human visual restoration while greatly reducing inference cost.

To validate the effectiveness of this lightweight configuration, we conduct ablation experiments (see Table \ref{abl3}), which demonstrate that the simplified design maintains strong enhancement capability with minimal computational overhead.

\subsubsection{\textbf{Visual Enhancement in the Technical Branch}}
The input to the technical branch differs significantly from that of the aesthetic branch. Instead of using globally resized frames, we directly sample patch blocks from the original resolution video frames, following the fragment-based strategy introduced in FastVQA \cite{wu2022fast}. These patches are assembled into fragments as input to the network. To better preserve fine-grained structural details, we select consecutive frames as input, minimizing visual artifacts caused by inter-frame jitter.
This design results in two key differences from the aesthetic branch: (1) the overall input appears more abstract, lacking complete semantic content, and (2) the temporal continuity between adjacent frames reduces inter-frame differences, making optical flow less informative. As a result, unlike the aesthetic branch, we do not apply the same visual enhancement network here. Instead, the technical branch focuses more on local content fidelity and fine-detail quality prediction, without relying on motion-based alignment mechanisms.

Inspired by RealBasicVSR \cite{chan2022investigating}, which demonstrates that simple residual blocks can effectively perform basic visual enhancement and image quality improvement, we design a lightweight enhancement module named CleanNet, composed entirely of residual blocks.
To compensate for the model's simplicity while improving enhancement effectiveness, we incorporate an iterative refinement strategy, as illustrated in Fig.~\ref{E_A_and_E_T}(b). Instead of enhancing the input in a single step, CleanNet progressively improves image quality over multiple iterations. We empirically fix the number of refinement steps to 3 during training, achieving a balance between visual quality and computational efficiency.

To validate the design choices for the technical branch, we conduct ablation studies (see Table \ref{abl3}) demonstrating the impact of CleanNet and its iterative mechanism on overall VQA performance. These results confirm that even without complex alignment mechanisms (e.g., optical flow), a residual-block-based enhancement approach can significantly benefit patch-level quality prediction.

\begin{figure*}[t]
    \centering
    \includegraphics[width=1\linewidth]{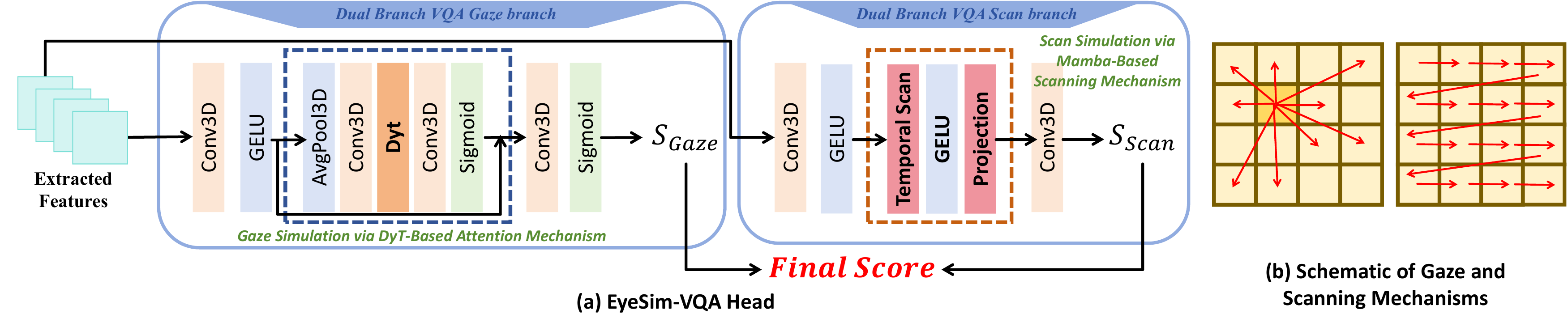}
    \caption{(a) Schematic illustration of the proposed EyeSim-VQA Head, which simulates the human visual mechanisms of gaze and scanning through a dual-branch architecture. (b) Visualization of the scanning mechanism using Mamba for horizontal feature sweeping (simulating human eye scanning), and the gaze mechanism based on attention computation, which models human fixation by estimating the relevance of each pixel to others.}
    \label{Dual_head}
\end{figure*}

\subsection{EyeSim-VQA Head: Modeling Human Gaze and Scanning Mechanisms} \label{sec_eyesim}

\subsubsection{\textbf{Gaze Branch: DyT-Based Attention Mechanism}}
To simulate the human eye’s gaze mechanism, we adopt a lightweight channel-wise attention strategy. To further improve performance and inference efficiency, we replace the standard LayerNorm \cite{xu2019understanding} layers with the DyT \cite{zhu2025transformers} architecture, which offers dynamic and learnable non-linearity. The specific formulation of DyT is provided below:
\begin{equation}
\mathrm{DyT}(\mathbf{x}) = \tanh(\alpha \cdot \mathbf{x}) \odot \mathbf{w} + \mathbf{b}
\end{equation}
The DyT module applies a dynamic non-linear transformation to the input feature tensor $\mathbf{x}$. Specifically, $\alpha$ is a learnable scalar parameter that adaptively scales the input prior to non-linearity. The $\tanh$ activation introduces smooth bounded non-linearity, promoting stable optimization. The resulting activation is then modulated by an element-wise learned scale parameter $\mathbf{w}$ and shifted by a learned bias term $\mathbf{b}$. The element-wise product $\odot$ allows channel-wise reweighting, enabling more expressive and adaptive transformations across the feature space.
Based on the DyT formulation, we design a lightweight attention mechanism to simulate the gaze behavior of the human eye, as illustrated in Fig. \ref{Dual_head}(a). The corresponding formulation is given below:
\begin{equation}
\left\{
\begin{aligned}
\mathbf{Q} &= \text{AvgPool}(\mathbf{Z}) \in \mathbb{R}^{B \times C \times 1 \times 1 \times 1} \\
\mathbf{K} &= \mathbf{Q} \\
\mathbf{V} &= \mathbf{Z} \in \mathbb{R}^{B \times C \times T \times H \times W} \\
\hat{\mathbf{A}} &= \phi \left( \text{DyT} \left( \mathbf{W}_1 \mathbf{Q} \right) \right) \in \mathbb{R}^{B \times C' \times 1 \times 1 \times 1} \\
\mathbf{A} &= \sigma \left( \mathbf{W}_2 \hat{\mathbf{A}} \right) \in \mathbb{R}^{B \times C \times 1 \times 1 \times 1} \\
\mathbf{O} &= \mathbf{A} \odot \mathbf{V} \in \mathbb{R}^{B \times C \times T \times H \times W}
\end{aligned}
\right.
\label{eq:se_dyt_attention}
\end{equation}
where $\mathbf{Z}$ denotes the compressed feature representation, $\phi(\cdot)$ is a non-linear activation function (e.g., GELU), $\sigma(\cdot)$ represents the Sigmoid function, and $\mathbf{W}_1$ and $\mathbf{W}_2$ are learnable $1{\times}1{\times}1$ convolutional kernels.
As illustrated in Fig. \ref{Dual_head}(b), this attention-based formulation computes the relevance between each pixel block and all others, effectively simulating the human eye’s gaze mechanism by selectively focusing on salient regions within the visual input.

\subsubsection{\textbf{Scan Branch: Mamba-Based Scanning Mechanism}}
We first recall the standard formulation of a discrete-time state space model (SSM) \cite{gu2023mamba}, which underpins the Mamba \cite{gu2023mamba,liu2024vision,zhang2024survey} architecture. As shown in Eq. \ref{eq2}, the system evolves over time through a hidden state $\mathbf{h}_t$ that is recursively updated based on the previous state and current input:
\begin{equation}
\left\{
\begin{aligned}
\mathbf{h}_{t} &= \mathbf{A} \mathbf{h}_{t-1} + \mathbf{B} \mathbf{u}_{t} \\
\mathbf{y}_{t} &= \mathbf{C} \mathbf{h}_{t} + \mathbf{D} \mathbf{u}_{t}
\end{aligned}
\right.
\label{eq2}
\end{equation}
Here, $\mathbf{A}, \mathbf{B}, \mathbf{C}, \mathbf{D}$ are learnable matrices that govern temporal dynamics and input-output transformation. This formulation enables long-range temporal modeling by maintaining a state trajectory that integrates both past memory and current observations. Mamba builds upon this principle by parameterizing these operators with efficient convolutional or kernel approximations to enable scalable sequence modeling.

To efficiently simulate this behavior in our temporal module, we design a lightweight alternative inspired by Mamba’s scan-then-project paradigm, as defined in Eq. \ref{eq3}:
\begin{equation}
\left\{
\begin{aligned}
\mathbf{Z}(t, h, w) &= \phi\left( \sum_{\tau=-1}^{1} \mathbf{W}_\tau \cdot \mathbf{X}(t+\tau, h, w) \right) \\
\mathbf{Y}(t, h, w) &= \mathbf{P} \cdot \mathbf{Z}(t, h, w)
\end{aligned}
\right.
\label{eq3}
\end{equation}
In this formulation, a 1D temporal convolution kernel $\{ \mathbf{W}_{-1}, \mathbf{W}_0, \mathbf{W}_{+1} \}$ aggregates features across a local temporal neighborhood to model short-term dynamics. The result is passed through a nonlinearity $\phi(\cdot)$, followed by a channel-wise projection $\mathbf{P}$, restoring the feature dimensionality. This mimics the state update and output projection process in Mamba while remaining highly efficient and compact, making it well-suited for VQA tasks under limited computational budgets.

The detailed architecture of the scan branch is shown in Fig. \ref{Dual_head}, where the pixel-wise sequential scanning driven by Mamba’s linear expansion effectively mimics the sweeping mechanism of human visual perception.

\begin{figure}[t]
    \centering
    \includegraphics[width=0.95\linewidth]{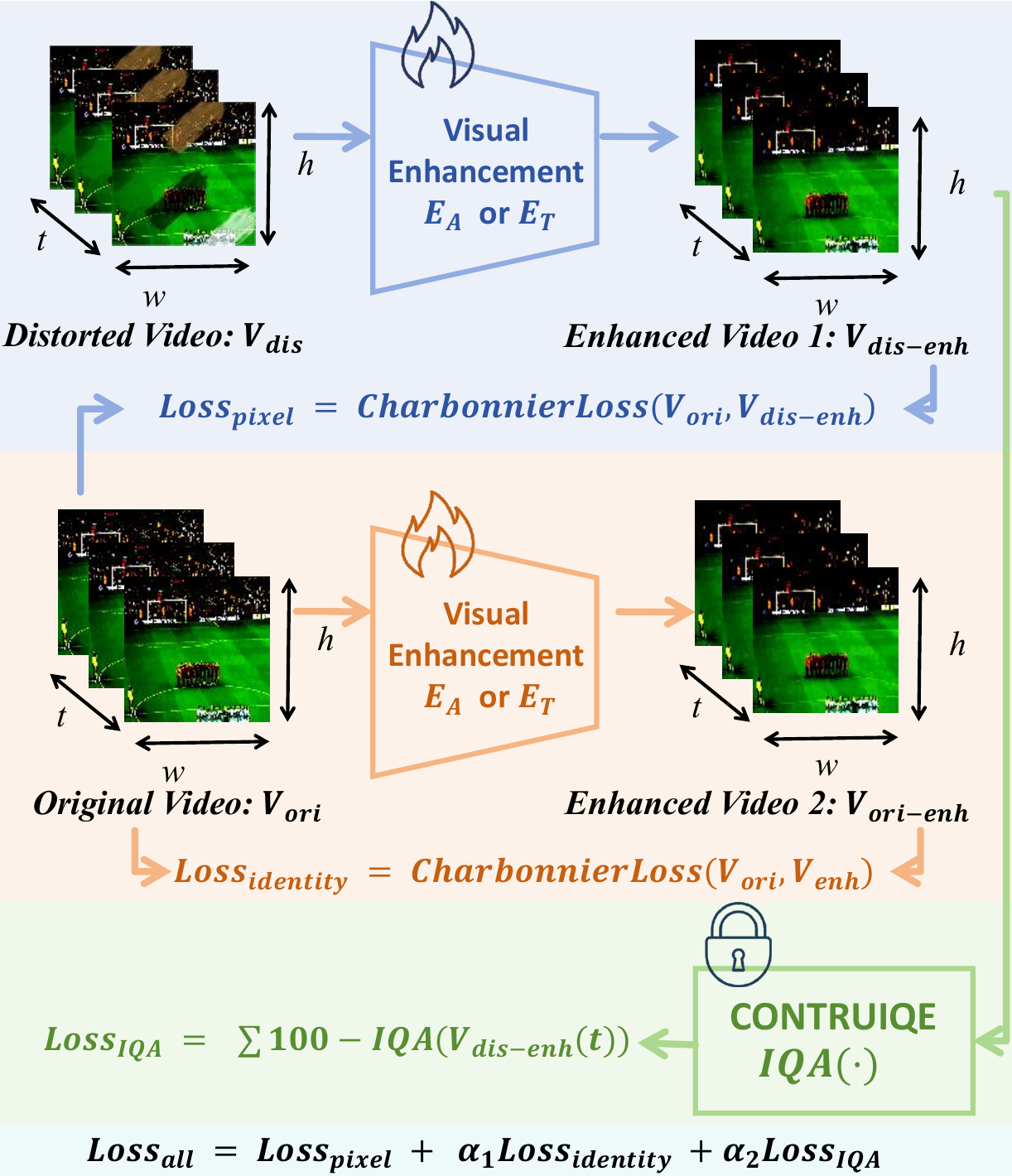}
    \caption{Overview of the training pipeline for visual enhancement.
The optimization is guided by a composite loss consisting of three key components: pixel loss, identity loss, and quality-aware supervision loss.}
    \label{loss}
\end{figure}

\subsection{Loss Function Design} \label{sec_loss}
To effectively train a visual enhancement network that simulates the self-repair capability of the human eye, we propose a tailored loss function. To facilitate supervised learning, we construct an artificial distortion dataset based on KoNViD-1k, following the methodology detailed in Sec \ref{shujuji}. The proposed loss function comprises three complementary components, as illustrated schematically in Fig. \ref{loss}.
The first component is a pixel-level reconstruction loss, implemented using the Charbonnier loss \cite{charbonnier1994two}, which is known for its robustness to outliers and smooth gradient properties. It is defined as:
\begin{equation}
\text{CharbonnierLoss}(x, y) = \sqrt{(x - y)^2 + \epsilon^2}
\label{eq:charbonnier}
\end{equation}
where $\epsilon$ is a small constant (typically set to $10^{-3}$) to ensure numerical stability. The specific pixel loss is shown below:
\begin{equation}
\mathcal{L}_{\text{pixel}} = \text{CharbonnierLoss}(\mathbf{V}_{\text{ori}}, \mathbf{V}_{\text{dis-enh}})
\label{eq:pixel_loss}
\end{equation}
where $V_{\text{ori}}$ denotes the original high-quality frame, $V_{\text{dis-enh}}$ is the enhanced distorted frame.
Following common practices in video super-resolution (VSR) tasks, we define the pixel-level loss $\mathcal{L}_{\text{pixel}}$ to ensure that the model is capable of restoring fine-grained details during enhancement. However, given that the input dataset contains both high- and low-quality samples, not all frames require enhancement. To prevent the model from unnecessarily altering already high-quality content, we introduce an identity preservation loss $\mathcal{L}_{\text{identity}}$. This loss acts as a regularization term, encouraging the network to retain the original content when no enhancement is needed. It is also formulated using the Charbonnier loss as:
\begin{equation}
\mathcal{L}_{\text{identity}} = \text{CharbonnierLoss}(\mathbf{V}_{\text{ori}}, \mathbf{V}_{\text{ori-enh}})
\label{eq:loss_identity}
\end{equation}
where $V_{\text{ori-enh}}$ denotes the corresponding enhanced output.
Finally, to better align the enhancement process with subjective human perception, we introduce a perceptual quality loss $\mathcal{L}_{\text{IQA}}$, guided by a pre-trained NR-IQA model—CONTRIQUE \cite{madhusudana2022image}. Specifically, we sample a set of frames from the enhanced video and compute their predicted quality scores using CONTRIQUE. These scores are then used to formulate a quality-based penalty, encouraging the network to generate outputs that are perceptually superior. The IQA loss is defined as:
\begin{equation}
\mathcal{L}_{\text{IQA}} = \sum_{t=1}^{T} \left( 100 - \text{IQA}(\mathbf{V}_{\text{dis-enh}}(t)) \right)
\label{eq:loss_iqa}
\end{equation}
where $V_{\text{dis-enh}}(t)$ denotes the enhanced frame at time $t$, and $\text{IQA}(\cdot)$ returns the predicted quality score in the range of [0, 100]. This formulation penalizes outputs with lower perceptual quality, guiding the enhancement network toward more human-aligned restoration.

Summing up all three components, the total loss used to supervise the training of the visual enhancement network is formulated as:
\begin{equation}
\mathcal{L}_{\text{total}} = \mathcal{L}_{\text{pixel}} + \alpha_1 \mathcal{L}_{\text{identity}} + \alpha_2 \mathcal{L}_{\text{IQA}}
\label{eq:loss_total}
\end{equation}
where $\alpha_1$ and $\alpha_2$ are hyperparameters controlling the relative contribution of the identity and IQA losses, respectively.

\begin{table*}[t]
\centering
\caption{Performance of existing SOTA methods. The
“N/A” means missing corresponding results in the original paper. The best and second-best results are bolded and underlined.}
\resizebox{1\textwidth}{!}{
\begin{tabular}{lcccccccccccc}
\toprule
\multirow{2}{*}{Method} & \multicolumn{2}{c}{\textbf{KVQ}} & \multicolumn{2}{c}{\textbf{KoNViD-1k}} & \multicolumn{2}{c}{\textbf{YouTube-UGC}} & \multicolumn{2}{c}{\textbf{LIVE-VQC}} & \multicolumn{2}{c}{\textbf{LSVQ test}} & \multicolumn{2}{c}{\textbf{LSVQ 1080p}} \\
\cmidrule(lr){4-5} \cmidrule(lr){6-7} \cmidrule(lr){2-3} \cmidrule(lr){8-9} \cmidrule(lr){10-11} \cmidrule(lr){12-13}
& SROCC & PLCC & SROCC & PLCC & SROCC & PLCC & SROCC & PLCC & SROCC & PLCC & SROCC & PLCC \\
\midrule
\multicolumn{13}{l}{\textit{Classical Approaches (based on handcraft features):}}\\
\hdashline
VIQE \cite{zheng2022completely}        & 0.221 & 0.397 & 0.628 & 0.638 & 0.513 & 0.476 & 0.659 & 0.694 & N/A & N/A & N/A & N/A \\
TLVQM \cite{korhonen2019two}       & 0.490 & 0.509 & 0.773 & 0.768 & 0.669 & 0.659 & 0.798 & 0.802 & 0.772 & 0.774 & 0.589 & 0.616 \\
RAPIQUE \cite{tu2021rapique}     & 0.740 & 0.717 & 0.803 & 0.817 & 0.759 & 0.768 & 0.754 & 0.786 & N/A & N/A & N/A & N/A \\
VIDEVAL \cite{tu2021ugc}     & 0.369 & 0.639 & 0.773 & 0.768 & 0.669 & 0.659 & 0.752 & 0.751 & 0.795 & 0.783 & 0.545 & 0.554 \\
\hdashline
\multicolumn{13}{l}{\textit{Deep Learning Approaches (based on deep neural network features):}}\\
\hdashline
VSFA \cite{li2019quality}        & 0.762 & 0.765 & 0.773 & 0.775 & 0.724 & 0.743 & 0.773 & 0.795 & 0.801 & 0.796 & 0.675 & 0.704 \\
GSTVQA \cite{chen2021learning}     & 0.786 & 0.781 & 0.814 & 0.825 & N/A  & N/A  & 0.788 & 0.796 & N/A & N/A & N/A & N/A \\
PVQ  \cite{ying2021patch}       & 0.794 & 0.801 & 0.791 & 0.786 & N/A  & N/A  & 0.827 & 0.837 & 0.827 & 0.828 & 0.711 & 0.739 \\
SimpleVQA \cite{sun2022deep}   & 0.840 & 0.847 & 0.856 & 0.860 & 0.847 & 0.856 & N/A  & N/A  & 0.867 & 0.861 & 0.764 & 0.803 \\
FastVQA \cite{wu2022fast}     & 0.832 & 0.834 & 0.891 & 0.892 & 0.855 & 0.852 & 0.849 & 0.862 & 0.876 & 0.877 & 0.779 & 0.814 \\
FasterVQA \cite{wu2023neighbourhood}     & N/A & N/A & 0.895 & 0.898 & 0.863 & 0.859 & 0.843 & 0.858 & 0.873 & 0.874 & 0.772 & 0.811 \\
Dover \cite{wu2022disentangling}      & 0.833 & 0.837 & 0.908 & 0.910 & 0.841 & 0.851 & 0.844 & 0.875 & 0.877 & 0.878 & 0.778 & 0.812 \\
PTM-VQA  \cite{yuan2024ptm}   & N/A & N/A & 0.856 & 0.857 & 0.857 & 0.811 & 0.819 & 0.819 & 0.811 & 0.863 & 0.735 & 0.781 \\
\hdashline
\multicolumn{13}{l}{\textit{CLIP-based or LLM-based Approaches:}}\\
\hdashline
CliF-VQA \cite{mi2024clif}     & N/A & N/A & 0.903 & 0.903 & 0.888 & 0.890 & \underline{0.866} & 0.878 & \underline{0.886} &\underline{ 0.887} & \underline{0.790} & \underline{0.832} \\
KSVQE \cite{lu2024kvq}       & \underline{0.867} & \underline{0.869} &\textbf{ 0.922} & \textbf{0.921} & 0.900 & \textbf{0.912} & 0.861 & \underline{0.883} & \underline{0.886} & \textbf{0.888} & \underline{0.790} & 0.823 \\
\midrule
\rowcolor{gray!20}
\textbf{EyeSim-VQA}   & \textbf{0.870} &\textbf{ 0.875} &\underline{0.919}  &\underline{0.918}  &\textbf{0.902}  &\underline{0.897}  &\textbf{0.886}  &\textbf{0.891}  &\textbf{0.888}  &0.885  &\textbf{0.807}  &\textbf{0.833} \\ 
\bottomrule
\end{tabular}
}
\label{main_ex1}
\end{table*}

\section{Experiments} \label{s4}
\subsection{Datasets and evaluation criteria}
\subsubsection{\textbf{Datasets}}
We evaluate our approach on five widely used VQA datasets: KVQ \cite{lu2024kvq}, KoNViD-1k \cite{hosu2017konstanz}, YouTube-UGC \cite{wang2019youtube}, LIVE-VQC \cite{sinno2018large}, and LSVQ \cite{ying2021patch}. For the first four datasets, we follow the partitioning protocol from our previous work by randomly splitting each dataset into training and testing sets at an 8:2 ratio. For LSVQ, we use the official training set for model training and report results on the official LSVQ-Test and LSVQ-1080 subsets. Our dataset usage and evaluation settings are aligned with mainstream VQA practices to ensure fair and consistent comparisons.
\subsubsection{\textbf{Evaluation Criteria}}
We adopt Pearson’s Linear Correlation Coefficient (PLCC) and Spearman’s Rank-Order Correlation Coefficient (SROCC) to evaluate prediction accuracy and monotonicity. Both metrics range from 0 to 1, where higher PLCC indicates better numerical alignment with MOS, and higher SROCC reflects more accurate ranking of sample quality.

\subsection{Implementation details} \label{shujuji}
All experiments are implemented in PyTorch and executed on an NVIDIA A6000 GPU. For the aesthetic branch, video frames are uniformly sampled with $N = 32$ and resized to 224×224. An Inflated ConvNeXt-Tiny \cite{liu2022convnet} model pretrained on AVA \cite{murray2012ava} serves as the backbone, while BasicVSR-mini is utilized for visual enhancement.
In the technical branch, input clips consist of 32 consecutive frames during training, and three clips are used during inference. From each frame, spatial patches of size $S_f = 32$ are extracted from a 7×7 grid. We employ Video Swin Transformer-Tiny \cite{liu2022video} with GRPB \cite{wu2022fast} as the backbone and introduce a CleanNet module based on residual blocks for enhancement. The loss function uses weights $\alpha_1 = 0.3$ and $\alpha_2 = 0.01$.

To train the visual enhancement network, we construct a synthetic distortion dataset based on KoNViD-1k by applying five common degradation types: Gaussian noise, motion blur, JPEG compression, salt-and-pepper noise, and mean blur. Distortion parameters are randomly sampled within appropriate ranges to simulate diverse real-world quality issues. These degradations are applied on-the-fly during training, with the original high-quality frames serving as supervision.
We train the enhancement model using the Adam optimizer with a learning rate of $1 \times 10^{-4}$, $\beta_1 = 0.9$, and $\beta_2 = 0.99$. A cosine annealing scheduler is employed with 400,000 total steps and a minimum learning rate of $1 \times 10^{-7}$.
When training the VQA network, we use the AdamW \cite{kingma2014adam,zhou2024towards} optimizer with a base learning rate of $1 \times 10^{-3}$ and a weight decay of 0.05. Following common practice in vision tasks, a smaller learning rate (scaled by a factor of 0.1) is applied to the backbone parameters. For learning rate scheduling, we adopt a warm-up strategy followed by cosine annealing.

\subsection{Comparison with SOTA methods}
\subsubsection{\textbf{Performance Evaluation on Individual Database}}
We compare our method against 13 representative approaches across five benchmark datasets. These include traditional methods (VIQE \cite{zheng2022completely}, TLVQM \cite{korhonen2019two}, RAPIQUE \cite{tu2021rapique}, VIDEVAL \cite{tu2021ugc}), deep learning-based models (VSFA \cite{li2019quality}, GSTVQA \cite{chen2021learning}, PVQ \cite{ying2021patch}, SimpleVQA \cite{sun2022deep}, FastVQA \cite{wu2022fast}, FasterVQA \cite{wu2023neighbourhood}, DOVER \cite{wu2022disentangling}, PTM-VQA \cite{yuan2024ptm}), and recent LLM-driven methods (CliF-VQA \cite{mi2024clif}, KSVQE \cite{lu2024kvq}).
As shown in Table \ref{main_ex1}, traditional methods struggle to deliver satisfactory performance under complex UGC-VQA scenarios, while deep learning-based approaches exhibit improved accuracy. More recent methods incorporating large language models (LLMs) further enhance performance by leveraging their semantic understanding capabilities. Among these, DOVER \cite{wu2022disentangling} achieves competitive results by modeling both global and local visual information, despite not utilizing LLMs. However, it lacks a mechanism explicitly aligned with human visual perception.
In contrast, our proposed model integrates a free-energy-guided visual self-repair mechanism that actively enhances video content during assessment, closely simulating the perceptual process of the human visual system. As a result, our method achieves consistently strong performance across benchmarks, and even outperforms LLM-based approaches on several datasets (e.g., KVQ, LIVE-VQC), demonstrating both effectiveness and superior perceptual alignment.

\subsubsection{\textbf{Performance Evaluation Cross Different Databases}}
To further assess the generalization capability of our model, we follow the experimental protocol outlined in the KVQ paper and conduct cross-dataset evaluations, as summarized in Table \ref{cross-kvq} and Table \ref{cross-other}. Specifically, we train the model on KVQ and test it on KoNViD-1k, YouTube-UGC, and LIVE-VQC, and vice versa—training on each of the latter datasets and evaluating on the KVQ test set.
The results show that our model consistently achieves strong performance across different datasets, demonstrating its robustness and generalizability. These findings further support the effectiveness of the proposed self-repair mechanism, which simulates human visual perception and contributes to improved cross-domain VQA performance.

\begin{table}[t]
\centering
\caption{Cross-dataset evaluations of ``other datasets $\rightarrow$ KVQ''. Each cell reports SROCC / PLCC on KVQ when trained on the indicated source dataset.}
\resizebox{0.48\textwidth}{!}{
\begin{tabular}{lccc}
\toprule
Test: KVQ & KoNViD-1k & YouTube-UGC & LIVE-VQC \\
\midrule
SimpleVQA \cite{sun2022deep} & 0.459 / 0.394 & 0.396 / 0.401 & 0.345 / 0.392 \\
FastVQA \cite{wu2022fast}  & 0.506 / 0.480 & 0.450 / 0.409 & 0.505 / 0.496 \\
KSVQE \cite{lu2024kvq}     & 0.528 / 0.504 & 0.499 / 0.412 & 0.539 / 0.533 \\
\midrule
EyeSim-VQA &\textbf{ 0.534 / 0.522}& \textbf{0.512 / 0.444}&\textbf{0.545 / 0.547} \\
 \bottomrule
\end{tabular}
}
\label{cross-kvq}
\end{table}

\begin{table}[t]
\centering
\caption{Cross-dataset evaluations of ``KVQ $\rightarrow$ other datasets''. Each cell reports SROCC / PLCC on the target dataset when trained on KVQ.}
\resizebox{0.48\textwidth}{!}{
\begin{tabular}{lccc}
\toprule
Train: KVQ & KoNViD-1k & YouTube-UGC & LIVE-VQC \\
\midrule
SimpleVQA \cite{sun2022deep}   & 0.475 / 0.481 & 0.675 / 0.674 & 0.528 / 0.521 \\
FastVQA \cite{wu2022fast}    & 0.641 / 0.654 & 0.645 / 0.676 & 0.614 / 0.675 \\
KSVQE \cite{lu2024kvq}      & \textbf{0.650 / 0.661} & 0.742 / 0.764 & 0.720 / \textbf{0.768} \\
\midrule
EyeSim-VQA &             0.645 / 0.654&            \textbf{ 0.751 / 0.771}&             \textbf{0.728} / 0.758\\
\bottomrule
\end{tabular}
}
\label{cross-other}
\end{table}

\subsubsection{\textbf{Performance Evaluation on the Ranking Pairs}}
Following the experimental setup in KVQ, we conduct Ranking Pairs experiments using 100 pairs of homologous and non-homologous videos provided by the official benchmark. The results are presented in Table \ref{kvq-ranking}. Our visual restoration-based VQA model demonstrates notable performance gains on non-homologous video pairs, where differences in quality are more pronounced. In such cases, the introduction of higher-level visual information leads to more accurate predictions. On homologous pairs, where quality differences are subtle, the performance gains are less significant. Nonetheless, our model achieves overall superior results across the full set of ranking pairs.

\begin{table}[t]
\centering
\caption{Performance on the ranking pairs in KVQ. There are 100 pairs, including 50 non-homogeneous and 50 homogeneous pairs.}
\resizebox{0.48\textwidth}{!}{
\begin{tabular}{lccc}
\toprule
Method & Non-Homogeneous & Homogeneous & All Pairs \\
\midrule
TLVQM \cite{korhonen2019two}& 0.56 & 0.64 & 0.60 \\
VIDEVAL \cite{tu2021ugc}   & 0.36 & 0.60 & 0.48 \\
\hdashline
VSFA \cite{li2019quality}          & 0.54 & 0.92 & 0.73 \\
GSTVQA \cite{chen2021learning}     & 0.58 & \textbf{0.98} & 0.78 \\
SimpleVQA \cite{sun2022deep} & 0.58 & 0.96 & 0.77 \\
FastVQA \cite{wu2022fast}   & 0.66 & 0.86 & 0.76 \\
Dover \cite{wu2022disentangling}      & 0.70 & 0.88 & 0.79 \\
KSVQE \cite{lu2024kvq}             & 0.76 & 0.86 & 0.81 \\
\midrule
EyeSim-VQA &\textbf{0.77}&0.88& \textbf{0.83}\\
\bottomrule
\end{tabular}
}
\label{kvq-ranking}
\end{table}


\subsection{Ablation studies}

\subsubsection{\textbf{Ablation on the individual components’ functionalities}}
We conduct ablation studies to evaluate the individual contributions of key components in our framework, as summarized in Table \ref{abl1}. Specifically, we assess the impact of the visual enhancement modules in the aesthetic and technical branches, as well as the proposed EyeSim-VQA prediction head.
Results indicate that replacing the original single-head predictor with our EyeSim head significantly improves the model’s ability to utilize extracted features, resulting in higher prediction accuracy. Additionally, the inclusion of visual enhancement modules—designed to simulate the human eye’s self-repair behavior—yields consistent performance gains across both branches.
The full model, which integrates all components, achieves the best overall performance. These findings confirm the effectiveness of each module and highlight the complementary roles they play in improving VQA accuracy.

\begin{figure*}[t]
    \centering
    \includegraphics[width=0.95\linewidth]{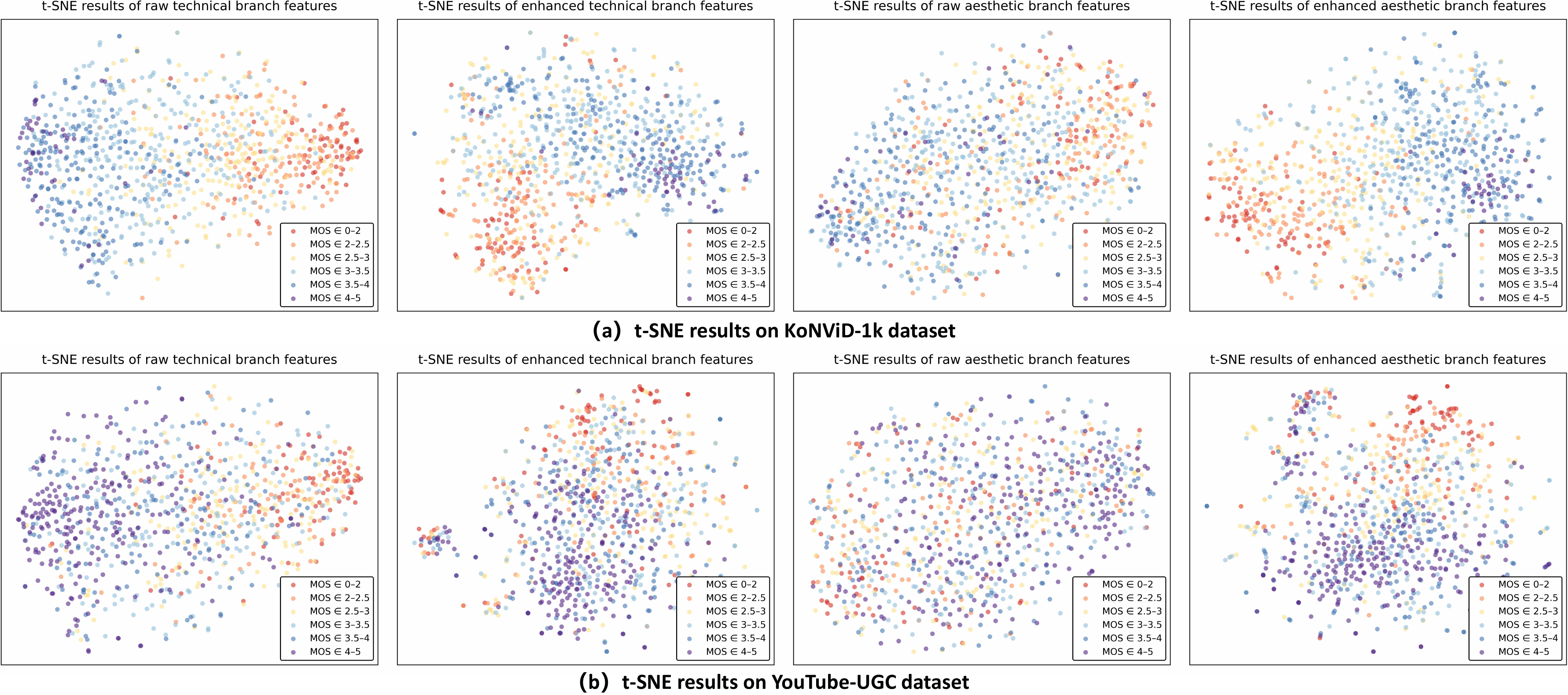}
    \caption{Visualization of clustering results using t-SNE \cite{van2009learning} on features extracted by the backbone networks from the aesthetic and technical branches. The inputs include both the original and quality-enhanced video frames. Videos from the KoNViD-1k and YouTube-UGC dataset are used, and the number of clusters is set to 6 based on the range of MOS values.}
    \label{tSNE}
\end{figure*}

\begin{table*}[t]
  \caption{Ablation study on individual components’ functionalities of the proposed model.}
    \centering
    \begin{tabular}{ccccccc}
    \toprule
    \multirow{2}{*}{Aesthetic Enhancement}&\multirow{2}{*}{Technical Enhancement}  & \multirow{2}{*}{EyeSim Head} & \multicolumn{2}{c}{\textbf{KoNViD-1k}}& \multicolumn{2}{c}{\textbf{YouTube-UGC}}\\
    \cmidrule(lr){4-5} \cmidrule(lr){6-7}
    &&&  SROCC $\uparrow$& PLCC $\uparrow$ & SROCC $\uparrow$&PLCC $\uparrow$ \\
   \midrule
    \xmark& \xmark&  \xmark&  0.908& 0.910&0.841& 0.851\\
    \xmark& \xmark& \cmark& 0.910&0.913&0.866&0.871\\
    \cmark&\xmark&  \cmark&  0.914&  0.915&0.886&0.879\\
   \xmark& \cmark&  \cmark&  0.915&  0.914&0.890&0.880\\
 \hdashline \rowcolor{gray!20}
   \cmark&\cmark& \cmark& \textbf{0.919}& \textbf{0.918}&\textbf{0.902}&\textbf{0.897}\\
 \bottomrule
    \end{tabular}
  
    \label{abl1}
\end{table*}

\subsubsection{\textbf{t-SNE Visualization of Visual Enhancement Effects}}
\begin{figure}[t]
    \centering
    \includegraphics[width=0.95\linewidth]{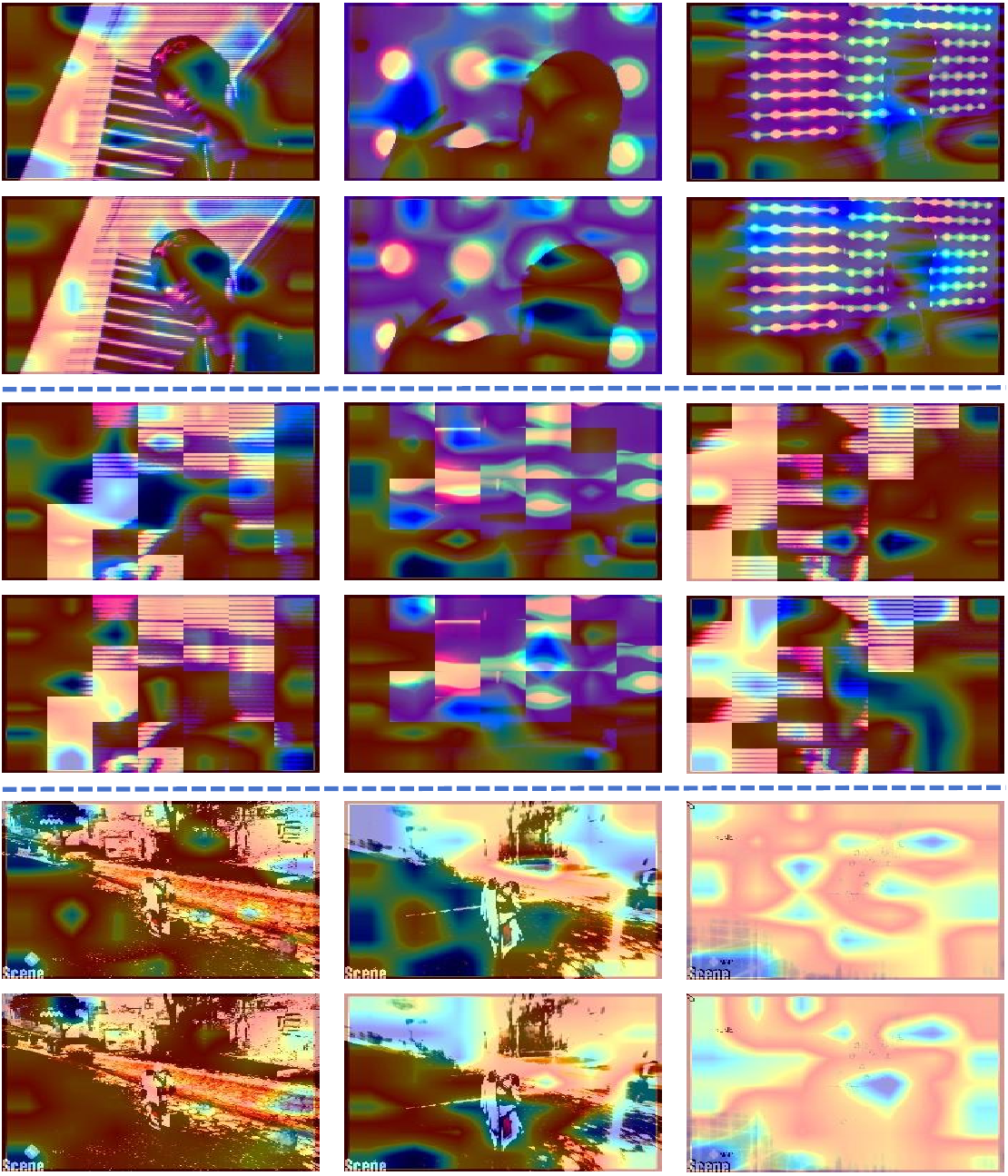}
    \caption{Visualization of the predicted attention maps from the proposed dual-branch VQA head.
We select three example images, including two aesthetic-dominant samples and one technical-dominant sample.
Since the final feature maps from the VQA head have low spatial resolution, we upsample and overlay them onto the original images to highlight the regions of focus for each branch.}
    \label{vis_score_weight}
\end{figure}

In addition to the quantitative results validating the effectiveness of our human-eye-inspired visual enhancement, we conduct qualitative experiments using t-SNE visualizations, as shown in Fig. \ref{vis_score_weight}. We evaluate our model on two benchmarks: KoNViD-1k and YouTube-UGC. For both the aesthetic and technical branches, we extract visual features from the original and enhanced inputs using their respective backbone networks. These features are then projected into a 2D space using t-SNE to examine their distribution with respect to MOS scores.

The results show that the original features already exhibit a certain degree of correlation with MOS scores. After visual enhancement, the features demonstrate a more compact and structured distribution, while maintaining alignment with quality levels. As illustrated in Fig. \ref{tSNE}(b), this effect is particularly pronounced in the aesthetic branch, where the enhanced features form distinct clusters corresponding to different quality grades.

This stronger distributional regularity provides intuitive evidence for the improved prediction performance, further validating the effectiveness of modeling the free-energy-guided self-repair mechanism in perceptual quality assessment.

\subsubsection{\textbf{Ablation on the EyeSim VQA Head Design}}
To validate the effectiveness of the proposed EyeSim-VQA prediction head, we conduct ablation experiments focusing on its core components, as summarized in Table \ref{abl2}. Specifically, we evaluate three key modules: (1) the dual-branch scoring and adaptive weighting mechanism, (2) the integration of DyT in the attention architecture, and (3) the Mamba-based sequential scanning mechanism.

Experimental results show that the dual-branch design significantly improves prediction accuracy by leveraging complementary information across gaze and scan pathways. Replacing the standard LayerNorm with the DyT architecture, as suggested in the original DyT paper, further enhances the attention mechanism, leading to measurable gains in performance. Additionally, the Mamba-based scanning mechanism, designed to mimic the human visual process of coarse-to-fine perception, contributes to improved perceptual alignment and score prediction.
When all components are combined, the model achieves the highest overall performance, confirming both the effectiveness of individual modules and the synergistic benefit of the full EyeSim-VQA head design.

\begin{table*}[t]
  \caption{Ablation study of the newly designed EysSim VQA prediction head to assess the contribution of each component.}
    \centering
    \begin{tabular}{ccc cccc}
    \toprule
    \multirow{2}{*}{Dual Branch}&\multirow{2}{*}{DyT}  & \multirow{2}{*}{Mamba-based Scan} & \multicolumn{2}{c}{\textbf{KoNViD-1k}}& \multicolumn{2}{c}{\textbf{YouTube-UGC}}\\
    \cmidrule(lr){4-5} \cmidrule(lr){6-7}
    &&&  SROCC $\uparrow$& PLCC $\uparrow$ & SROCC $\uparrow$&PLCC $\uparrow$ \\
   \midrule
    \xmark& \xmark&  \xmark&  0.910&   0.911& 0.887&0.874\\
    \cmark& \xmark& \xmark& 0.915& 0.914& 0.895&0.888\\
    \cmark&\cmark&  \xmark&  0.917&   0.914& 0.898&0.891\\
   \cmark& \xmark&  \cmark&  0.916&   0.915& 0.897&0.890\\
   \hdashline \rowcolor{gray!20}
   \cmark&\cmark& \cmark& \textbf{0.919}&  \textbf{0.918}& \textbf{0.902}&\textbf{0.897}\\
 \bottomrule
    \end{tabular}
  
    \label{abl2}
\end{table*}

\subsubsection{\textbf{Feature Map Visualization of the Dual-Branch Predictor Head}}
To further evaluate the rationality and effectiveness of the proposed dual-branch predictor head, we visualize the feature maps generated by each branch, as shown in Fig. \ref{vis_score_weight}. Since the input features to the VQA head have already undergone spatial compression via the backbone, the resulting feature maps are relatively small. To improve interpretability, we overlay these maps onto the original input frame, enabling intuitive analysis of region-specific sensitivity.
As shown in Fig. \ref{vis_score_weight}, the two branches exhibit distinct and complementary attention patterns, each emphasizing different spatial regions of the same frame. When combined, their responses lead to more balanced and semantically consistent predictions. This behavior is consistently observed across both aesthetic and technical inputs, providing qualitative support for the structural validity and perceptual alignment of our dual-branch design.

\subsubsection{\textbf{Ablation on Visual Enhancement Designs in Aesthetic and Technical Branches}}
We further conduct ablation studies on the design and selection of visual enhancement modules for both the aesthetic and technical branches, as summarized in Table \ref{abl3}. We evaluate three representative VSR models, including both reference-based and blind approaches. Among them, CleanNet is a lightweight enhancement module designed with simple residual blocks, while BasicVSR-mini is a significantly simplified version of BasicVSR \cite{chan2021basicvsr}, with reduced residual layers and feature channels.

In the technical branch, we observe that traditional VSR models relying on optical flow and adjacent frame information perform suboptimally. This is likely due to the use of small spatial patches with limited global context, and the fact that the input consists of 32 consecutive frames, where motion-based features are less effective. In contrast, the simpler CleanNet yields better performance, likely because it focuses on local structural enhancement without relying on temporal alignment.
For the aesthetic branch, performance differences across VSR models are less significant. To balance performance and efficiency, we adopt BasicVSR-mini as the enhancement module for this branch, as it offers a good trade-off between model complexity and prediction accuracy, while maintaining competitive results.

\begin{figure}[t]
    \centering
    \includegraphics[width=0.95\linewidth]{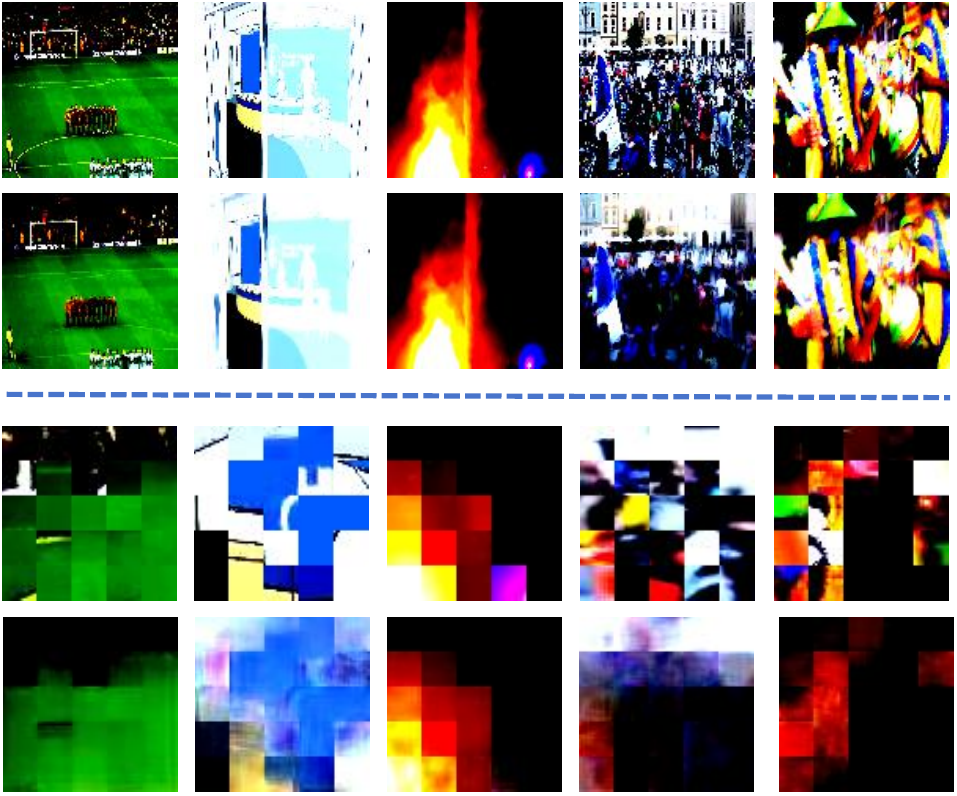}
    \caption{Visualization of video quality enhancement results from the Aesthetic and Technical branches. Each column corresponds to the same frame from a given video. From top to bottom: (1) original frame resized to a fixed resolution, (2) enhanced result after quality restoration, (3) reconstructed frame from extracted patches, and (4) enhanced patch-based result. All samples are selected from the KoNViD-1k dataset.}
    \label{vis_enhancement}
\end{figure}

\subsubsection{\textbf{Visualization of the Human Eye's self-repair Process}}
To provide a more intuitive understanding of how our model simulates the human eye’s self-repair mechanism, we visualize the outputs after the dual-branch visual enhancement process, as shown in Fig. \ref{vis_enhancement}.
In the aesthetic branch, the enhancement module effectively suppresses noise across the entire image. For example, in the second column of the first two rows, visible noise speckles are removed, resulting in cleaner and more visually pleasing outputs. This denoising, however, comes with a minor trade-off: some fine-grained textures are smoothed, leading to a slightly softer overall appearance.
In contrast, the technical branch operates on spatially partitioned image patches, which initially appear visually fragmented and structurally disjoint. After enhancement, the boundaries between patches become significantly less pronounced, producing a more coherent and unified visual output. This transformation induces a larger perceptual shift compared to the aesthetic branch, highlighting the enhancement module’s ability to restore spatial continuity and structural integrity in highly degraded or patch-based inputs.

\begin{table}[t]
\caption{Ablation study on different visual enhancement networks applied to the two branches of the proposed model. Results are tested on the KoNViD-1k dataset.}
\centering
\begin{tabular}{cccc}
    \toprule
    \multicolumn{2}{c}{Visual Enhancement}&  \multirow{2}{*}{SROCC $\uparrow$}& \multirow{2}{*}{PLCC $\uparrow$}\\
    \cmidrule(lr){0-1}
    $E_T$& $E_A$& &\\
    \midrule
    BasicVSR \cite{chan2021basicvsr}&  BasicVSR 
&  0.904&0.910\\
    BasicVSR++ \cite{chan2022basicvsr++}& BasicVSR++ 
& 0.910&0.908\\
    CleanNet&  RealBasicVSR \cite{chan2022investigating}& 0.917& 0.918\\
    \hdashline \rowcolor{gray!20}
    CleanNet&  BasicVSR-mini&  \textbf{0.919}&\textbf{0.918}\\
    \bottomrule
\end{tabular}
\label{abl3}
\end{table}

\subsubsection{\textbf{Ablation on Visual Enhancement Feature Embedding Strategies}}
After obtaining the visually enhanced data, we conduct ablation experiments to explore different strategies for embedding the enhanced features into the model, as shown in Table \ref{abl4}. Specifically, we evaluate four commonly used fusion methods for incorporating high-level visual information.

The first approach applies direct image-level fusion by combining the enhanced frames with the original inputs prior to feature extraction. However, this disrupts low-level features critical for accurate VQA predictions, leading to poor performance.
The second approach employs a residual fusion strategy, treating the enhanced output as a residual signal added to the original input. While this reduces the dominance of raw features, it also limits the use of high-level information and may disturb spatial structures, resulting in suboptimal outcomes.
The third method follows an adapter-based design, injecting enhanced features into intermediate layers of the backbone. Although this introduces additional cues, it alters the network architecture and interferes with pre-trained weights, degrading performance.
In contrast, a feature concatenation strategy—where enhanced features are concatenated with the original inputs and jointly processed by the backbone—achieves the best performance. This method preserves original feature integrity, effectively integrates enhancement cues, and maintains compatibility with pre-trained backbone parameters.

\begin{table}[t]
  \caption{Ablation study on the enhanced visual feature embedding strategies. Results are evaluated on the KoNViD-1k dataset.}
    \centering
    \begin{tabular}{lcc}
    \toprule
              \textit{Enhanced Visual Feature Embedding Strategies}&  SROCC $\uparrow$& PLCC $\uparrow$\\
           \midrule
            Image Fusion&  0.875&  0.884\\
            Residual Image Fusion& 0.901&0.897\\
 Adapter Integration into  $Backbone$&  0.905&  0.904\\
 \hdashline \rowcolor{gray!20}
           \textbf{Simple Concatenation}&  \textbf{0.919}&  \textbf{0.918}\\
    \bottomrule
    \end{tabular}
  
    \label{abl4}
\end{table}

\subsubsection{\textbf{Ablation on the Utilization Ratio of Enhanced Visual Frames}}
After obtaining the enhanced high-level visual features, we conduct ablation studies to investigate how they should be incorporated within the input sequence, as shown in Table \ref{abl5}. Specifically, we evaluate configurations using 16 and 32 original frames, as well as variants where a portion of the frames is replaced by visually enhanced frames via concatenation.
The results suggest that the inherent temporal structure—namely, the native frame content and frame rate—remains the most critical factor for VQA performance. Enhanced frames are most effective when used as complementary guidance rather than as replacements for the original input. Notably, for 32-frame inputs, replacing approximately one-third of the frames with enhanced versions yields the best results, striking an effective balance between structural fidelity and perceptual enhancement.

\begin{table}[t]
  \caption{Ablation study on the number of enhanced video frames selected for quality prediction. Results are tested on the KoNViD-1k dataset.}
    \centering
\begin{tabular}{cccc}
    \toprule
     Input Frame&Concatenated Frame&  SROCC $\uparrow$& PLCC $\uparrow$\\
    \midrule
    16&5&  0.910&  0.911\\
    16&10& 0.909&0.904\\
    32&5&  0.914&  0.916\\
    \hdashline \rowcolor{gray!20}
    32& 10& \textbf{0.919}&\textbf{0.918}\\
    \hdashline
    32&20&  0.911&  0.915\\
    \bottomrule
\end{tabular}

    \label{abl5}
\end{table}

\subsubsection{\textbf{Analysis of DyT Integration in Backbone and Prediction Head}}
In \cite{zhu2025transformers}, Zhu \textit{et al.} demonstrate that DyT can effectively replace LayerNorm in Transformer-based architectures for vision tasks such as image classification and segmentation. To evaluate its applicability in the VQA setting, we perform ablation experiments, as summarized in Table \ref{abl6}. Specifically, we replace the activation and normalization layers in both branches of the backbone network with DyT prior to training. However, this configuration yields suboptimal performance.
We hypothesize that this degradation stems from substantial alterations to the backbone architecture, which interfere with the pre-trained weights and reduce transferability to the VQA task. In contrast, applying DyT only within the VQA prediction head—which is trained from scratch—yields clear performance improvements. This is likely due to the predictor head’s smaller parameter footprint, which avoids disrupting the learned feature representations in the backbone.
We further speculate that DyT may demonstrate improved performance in backbone architectures when trained on larger-scale VQA datasets, where sufficient data allows better adaptation to such architectural modifications.

\begin{table}[t]
  \caption{Ablation study on the replacement of the DyT structure within the proposed framework. Results are tested on the KoNViD-1k dataset.}
    \centering
 \resizebox{0.48\textwidth}{!}{
    \begin{tabular}{ccccc}
    \toprule
             Backbone $M_A$&Backbone $M_T$&VQA head&  SROCC $\uparrow$& PLCC $\uparrow$\\
           \midrule
            \xmark&\xmark&\xmark&  0.916&  0.915\\
 \cmark&\xmark&\xmark& 0.899&0.900\\
            \xmark&\cmark&\xmark&  0.887&  0.894\\
            \hdashline \rowcolor{gray!20}
 \xmark&\xmark& \cmark& \textbf{0.919}&\textbf{0.918}\\
 \hdashline 
           \cmark&\cmark&\cmark&  0.875&  0.890\\
    \bottomrule
    \end{tabular}
    }
  
    \label{abl6}
\end{table}

\subsubsection{\textbf{Ablation on the Modular Design of the Loss Function for Visual Enhancement}}
In training the visual enhancement module, we design a composite loss function consisting of three components, and conduct ablation experiments to evaluate the individual contribution of each, as shown in Table \ref{abl7}.
The first component is the pixel-wise loss ($\mathcal{L}_{\text{pixel}}$), which serves as the foundation by ensuring basic fidelity and preserving low-level details during restoration. Since the low-quality data is synthetically generated, we also introduce an identity loss ($\mathcal{L}_{\text{identity}}$) to prevent the model from unnecessarily modifying already high-quality regions. The third component incorporates guidance from a pre-trained IQA model, which aligns the learning process with human perceptual judgments. However, we observe that using the IQA loss ($\mathcal{L}_{\text{IQA}}$) alone—without pixel-level supervision—leads to suboptimal results due to its limited sensitivity to fine-grained spatial details.
When all three losses are combined, the model benefits from both structural accuracy and perceptual alignment, leading to the best overall performance.

\begin{table}[t]
  \caption{Ablation study on the design of loss function hyperparameters. Results are tested on the KoNViD-1k dataset.}
    \centering
    \begin{tabular}{ccccc}
    \toprule
               $\mathcal{L}_{\text{pixel}}$&$\mathcal{L}_{\text{identity}}$&$\mathcal{L}_{\text{IQA}}$&  SROCC $\uparrow$& PLCC $\uparrow$\\
           \midrule
             \cmark&\xmark&\xmark&  0.910&  0.911\\
             \xmark&\xmark&\cmark& 0.899&0.905\\
  \cmark&\cmark&\xmark&  0.912&  0.913\\
  \cmark&\xmark&\cmark& 0.916&0.914\\
   \hdashline \rowcolor{gray!20}
            \cmark&\cmark&\cmark&  \textbf{0.919}&  \textbf{0.918}\\
    \bottomrule
    \end{tabular}
  
    \label{abl7}
\end{table}

\begin{table}[t]
  \caption{Ablation study on the design of loss function hyperparameters. Results are tested on the KoNViD-1k dataset.}
    \centering
    \begin{tabular}{cccc}
    \toprule
              $\alpha_1$&$\alpha_2$&  SROCC $\uparrow$& PLCC $\uparrow$\\
           \midrule
            0.1&0.01&  0.912&  0.910\\
            \hdashline \rowcolor{gray!20}
 0.3&0.01& \textbf{0.919}&\textbf{0.918}\\
 \hdashline
            0.3&0.03&  0.915&  0.916\\
 0.5&0.03& 0.914&0.911\\
           1&0.1&  0.900&  0.894\\
    \bottomrule
    \end{tabular}
  
    \label{abl8}
\end{table}

\subsubsection{\textbf{Ablation on Loss Function Hyperparameter Settings}}
The overall loss function consists of three components, with two hyperparameters used to balance their contributions. We conduct ablation experiments to investigate the impact of these hyperparameters, as shown in Table \ref{abl8}.
Among the three losses, the pixel-level loss ($\mathcal{L}_{\text{pixel}}$) serves as the primary supervisory signal, ensuring that the model restores basic visual fidelity. The identity loss ($\mathcal{L}_{\text{identity}}$) acts as a regularizer to prevent the model from unnecessarily altering already high-quality regions, though it does not directly guide the restoration process. The IQA loss ($\mathcal{L}_{\text{IQA}}$) is derived from the CONTRIQUE \cite{madhusudana2022image} model, which outputs quality scores in the range of 0–100. To ensure compatibility with the other loss components, this term is rescaled to match their magnitude. As the IQA loss is less sensitive to low-level pixel information, its corresponding weight ($\alpha_2$) should be kept relatively small. Through extensive experiments, we find that setting $\alpha_1=0.3$ and $\alpha_2=0.01$ yields the best performance, striking a balance between pixel-level accuracy, structural consistency, and perceptual quality.

\section{Limitation and Future Work} \label{s5}
While our framework effectively simulates key aspects of human visual perception—namely, the self-repair mechanism through dual-branch visual enhancement and the scan-and-gaze behavior through a biologically inspired prediction head—several limitations remain, suggesting promising directions for future exploration.

\textbf{First}, the current approach requires pre-training a dedicated visual enhancement module before VQA model training. This adds complexity compared to conventional end-to-end frameworks and may affect deployment efficiency in real-world scenarios.
\textbf{Second}, although LLMs have demonstrated strong semantic reasoning capabilities in VQA tasks, our framework does not yet incorporate LLM-based knowledge. Exploring how to integrate human-eye-inspired visual restoration with LLMs for unified perception and reasoning remains an important future direction.
\textbf{Third}, the enhancement modules in our framework are built upon a limited set of classical VSR architectures. There is substantial potential to explore more advanced or task-specific enhancement backbones that better align with the VQA objective. Our current design primarily serves as a proof of concept for modeling visual self-repair mechanisms; each component may benefit from further refinement as more powerful models become available.

In future work, we plan to address these limitations by investigating joint training strategies, incorporating LLMs, and adopting SOTA enhancement networks to further improve both restoration quality and VQA performance.

\section{Conclusion} \label{s6}
In this work, we propose EyeSim-VQA, a novel VQA framework inspired by the free-energy-guided self-repair mechanism of the human visual system. The proposed visual enhancement module mimics the eye's ability to restore degraded vision, while the dual-branch prediction head simulates human perception through sweeping and fixational gaze modeling. Extensive experiments on multiple VQA benchmarks demonstrate that our method achieves SOTA performance. Ablation studies further validate the effectiveness and rationality of each component. We believe that modeling human visual perception from both enhancement and attention perspectives opens new avenues for advancing VQA and related fields.

\bibliographystyle{IEEEtran}
\bibliography{reference}

\end{document}